\pdfoutput=1

\documentclass[11pt]{article}

\usepackage[]{acl}
\usepackage{algpseudocode,algorithm}
\usepackage{pgfplots}
\usepgfplotslibrary{fillbetween}

\usepackage{times}
\usepackage{latexsym}
\usepackage{amsmath}
\usepackage{amssymb} 
\usepackage[T1]{fontenc}
\usepackage{multirow}

\usepackage{booktabs} 
\usepackage{threeparttable}
\usepackage{filecontents}
\usepackage{hyperref}

\makeatletter
\newcommand{\printfnsymbol}[1]{%
  \textsuperscript{\@fnsymbol{#1}}%
}
\makeatother

\usepackage[utf8]{inputenc}
\usepackage{subcaption}
\usepackage{tikz}
\usetikzlibrary{shapes,positioning,calc, fit}
\tikzset{var/.style={draw,circle,fill=white,inner sep=1.5pt,minimum size=8pt}}
\tikzset{ext/.style={var,fill=black,text=white}}
\tikzset{fac/.style={draw,rectangle}}
\tikzset{subgraph/.style={draw,cloud}}
\tikzset{every picture/.style={baseline=-2.5pt}}
\tikzset{every label/.style={font={\footnotesize}}}
\tikzset{plate/.style={draw,rectangle,rounded corners}}
\tikzset{
  big arrow/.style={
    decoration={markings,mark=at position 1 with {\arrow[scale=4,#1]{>}}},
    postaction={decorate},
    shorten >=0.4pt},
  big arrow/.default=blue}
\tikzstyle{arrow} = [thick,->,>=stealth]

\usepackage{bm}
\newcommand{\vect}[1]{\boldsymbol{\mathbf{#1}}}

\usepackage{xcolor}
\usepackage{booktabs}
\usepackage{enumitem}
\newlist{compactenum}{enumerate}{1} 
\setlist[compactenum]{label=\arabic*.,ref=\arabic*,nosep,parsep=\parskip,leftmargin=2em}
\newlist{compactitem}{itemize}{3} 
\setlist[compactitem]{label=\textbullet,nosep,parsep=\parskip,leftmargin=2em}
\newcommand{\xlongrightarrow}[1]{\stackrel{#1}{\longrightarrow}}

\setlist[itemize]{noitemsep}

\newcommand{\rv}[1]{\mathbf{#1}}
\newcommand{\nt}[1]{\mathsf{#1}} 

\usepackage{microtype}

\title{Dynamic Programming in Rank Space: Scaling Structured Inference with Low-Rank HMMs and PCFGs}

\author{Songlin Yang\thanks{\; Equal contributions: Songlin Yang formulated the idea and Wei Liu conducted the experiments.  $^\dagger$ Corresponding author.} , Wei Liu\printfnsymbol{1}, Kewei Tu$^{\dagger}$\\
  School of Information Science and Technology, ShanghaiTech University \\
    Shanghai Engineering Research Center of Intelligent Vision and Imaging\\ 
    {\tt \{yangsl,liuwei4,tukw\}@shanghaitech.edu.cn}\\
 }

\begin{document}
\maketitle
\begin{abstract}
Hidden Markov Models (HMMs) and Probabilistic Context-Free Grammars (PCFGs) are widely used structured models, both of which can be represented as factor graph grammars (FGGs), a powerful formalism capable of describing a wide range of models. Recent research found it beneficial to use large state spaces for HMMs and PCFGs. However, inference with large state spaces is computationally demanding, especially for PCFGs. To tackle this challenge, we leverage tensor rank decomposition (aka.\ CPD) to decrease inference computational complexities for a subset of FGGs subsuming HMMs and PCFGs. We apply CPD on the factors of an FGG and then construct a new FGG defined in the rank space. Inference with the new FGG produces the same result but has a lower time complexity when the rank size is smaller than the state size. We conduct experiments on HMM language modeling and unsupervised PCFG parsing, showing better performance than previous work. Our code is publicly available at \url{https://github.com/VPeterV/RankSpace-Models}.


\end{abstract}

\section{Introduction}
Hidden Markov Models (HMMs) and Probabilistic Context-Free Grammars (PCFGs) are widely used structured models in natural language processing. They can both be represented as factor graph grammars (FGGs) \cite{DBLP:conf/nips/0001R20}, which are a powerful tool to describe a wide range of models, allowing exact and tractable inference in most situations of interest.

Over-parameterization has been shown beneficial in facilitating optimization of deep networks \cite{DBLP:conf/icml/AroraCH18, DBLP:conf/nips/XuHM18, DBLP:conf/iclr/DuZPS19}. \citet{DBLP:conf/icml/BuhaiHKRS20} found that over-parameterization is also helpful in learning latent variable models by increasing the number of hidden states. \citet{DBLP:conf/icml/BuhaiHKRS20} argued that it is important to study over-parameterization in structured settings because structured latent variable models are more suitable to model real-word phenomena which exhibit complex dependencies. HMMs and PCFGs are typical structured latent variable models, and recently researchers have found it beneficial to use large state spaces for HMMs and PCFGs \cite{DBLP:journals/corr/abs-1905-00507, chiu-rush-2020-scaling, yang-etal-2021-pcfgs, chiu2021low}.
However, structured inference with large state spaces is computationally demanding, especially for PCFGs, pushing researchers to develop methods to decrease the computational complexities. 
 \citet{chiu-rush-2020-scaling} propose a neural VL-HMM with $2^{15}$ states for language modeling, narrowing down the performance gap between HMMs and LSTMs. 
  They follow \citet{DBLP:journals/corr/abs-1905-00507} to impose a strong sparsity constraint (i.e., each hidden state can only generate a small subset of terminal symbols) to decrease the time complexity of the forward algorithm, thus requiring pre-clustering of terminal symbols.
\citet{yang-etal-2021-pcfgs} use a large state space for neural PCFG induction and achieve superior unsupervised constituency parsing performance. They follow \citet{cohen-etal-2013-approximate} to use tensor rank decomposition (aka.\ canonical-polyadic decomposition (CPD) \cite{DBLP:journals/corr/abs-1711-10781}) to decrease the computational complexity of the inside algorithm, but only scale the state size from tens to hundreds because the resulting complexity is still high. \citet{chiu2021low} use tensor matricization and low-rank matrix decomposition to accelerate structured inference on chain and tree structure models. However, their method has an even higher complexity than \citet{yang-etal-2021-pcfgs} on PCFGs. Recently, \citet{fu2021scaling} propose a family of randomized dynamic programming algorithms to scale structured models to tens of thousands of states, which is orthogonal to the aforementioned low-rank-based approaches as the former performs approximate inference whereas the latter perform exact inference. 



In this work, we propose a new low-rank-based approach to scale structured inference, which can be described by FGG notations intuitively.
We first provide an intuitive and unifying perspective toward the work of \citet{yang-etal-2021-pcfgs} and \citet{chiu2021low}, showing that their low-rank decomposition-based models can be viewed as decomposing large factors in an FGG---e.g., the binary rule probability tensor in PCFGs--- into several smaller factors connected by new ``rank'' nodes.  Then we target at a subset of FGGs---which we refer to as B-FGGs---subsuming all models considered by \citet{chiu2021low}, whereby the inference algorithms can be formulated via 
B-graphs \cite{DBLP:journals/dam/GalloLP93, DBLP:conf/iwpt/KleinM01}. We propose a novel framework to support a family of inference algorithms in the rank space for B-FGGs. Within the framework, we apply CPD on the factors of a B-FGG and then construct a new B-FGG defined in the rank space by marginalizing all the state nodes. Inference with the new B-FGG has the same result and a lower time complexity if the rank size is smaller than the state size.

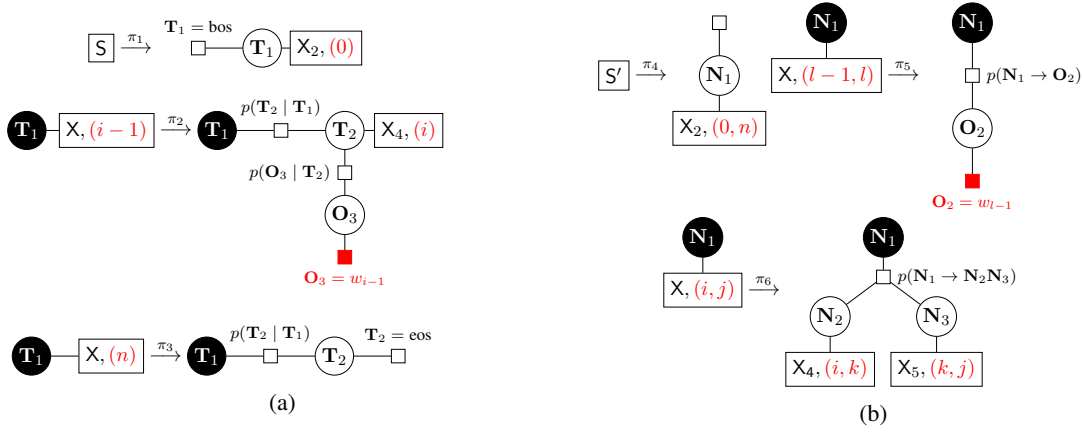
\begin{figure*}[tb!]
	\begin{subfigure}{0.48\linewidth}
	\scalebox{0.7}{
\begin{tabular}{c}
\begin{tikzpicture} 
\node[fac] at (0,0) { $\nt{S}$ }; 
\end{tikzpicture} 
$\xlongrightarrow{\pi_1}$
\begin{tikzpicture}[x=1.2cm,y=0.8cm]
\node[var] (t) at (0,0) {$\rv{T}_1$}; 
\node[fac,label=above:{$\rv{T}_1 = \text{bos}$}] at (-1,0) {} edge (t); 
\node[fac] at (1,0) {$\nt{X}_2, \textcolor{red}{(0)}$} edge (t); 
\end{tikzpicture} 
\\
\\
\begin{tikzpicture} 
\node[ext] (x) at (-1,0) {$\rv{T}_1$}; \node[fac] at (0.5,0) { $\nt{X}, \textcolor{red}{(i-1)}$} edge (x); 
\end{tikzpicture} 
$\xlongrightarrow{\pi_2}$
\begin{tikzpicture}[x=1.2cm,y=0.8cm] 
\node[ext] (x) at (-2,0) {$\rv{T}_1$}; 
\node[var] (t) at (0,0) {$\rv{T}_2$}; 
\node[var] (w) at (0,-2) {$\rv{O}_3$};
\node[fac,label=above:{$p(\rv{T}_2\mid \rv{T}_1)$}] at (-1,0) {} edge (x) edge (t); 
\node[fac,label=left:{$p(\rv{O}_3 \mid \rv{T}_2)$}] at (0,-1) {} edge (t) edge (w);
\node[fac] at (1,0) {$\nt{X}_4,  \textcolor{red}{(i)}$ } edge (t); 
\node[fac, red, fill, label=below:{\textcolor{red}{$\rv{O}_3=w_{i-1}$}}] at (0,-3) {} edge (w);
\end{tikzpicture} 
\\
\\
\begin{tikzpicture} 
\node[ext] (x) at (-1,0) {$\rv{T}_1$}; \node[fac] at (0.5,0) { $\nt{X}, \textcolor{red}{(n)}$ } edge (x); 
\end{tikzpicture} 
$\xlongrightarrow{\pi_3}$
\begin{tikzpicture}[x=1.2cm,y=0.8cm] 
\node[ext] (t1) at (-2,0) {$\rv{T}_1$}; 
\node[var] (t2) at (0,0) {$\rv{T}_2$}; 
\node[fac,label=above:{$p(\rv{T}_2 \mid \rv{T}_1)$}] at (-1,0) {} edge (t1) edge (t2);
\node[fac,label=above:{$\rv{T}_2 = \text{eos}$}] at (1,0) {} edge (t2); 
\end{tikzpicture}
\end{tabular}
}
\caption{}
\end{subfigure}
\begin{subfigure}{0.48\linewidth}
\scalebox{0.7}{
\begin{tabular}{c}
\begin{tikzpicture} 
\node[fac] at (0,0) { $\nt{S'}$ }; 
\end{tikzpicture} 
$\xlongrightarrow{\pi_4}$
\begin{tikzpicture} 
\node[var] (n) at (0,0) { $\rv{N}_1$ }; 
\node[fac,label=right:{}] at (0,1) {} edge (n);
\node[fac] at (0,-1) {$\nt{X}_2, \textcolor{red}{(0, n)}$} edge (n); 
\end{tikzpicture} 
\begin{tikzpicture} 
\node[ext](x) at (0,1) {$\rv{N}_1$};
\node[fac] at (0,0) { $\nt{X}, \textcolor{red}{(l-1, l)}$} edge (x); 
\end{tikzpicture} 
$\xlongrightarrow{\pi_5}$
\begin{tikzpicture} 
\node[ext] (n) at (0,1) { $\rv{N}_1$ }; 
\node[var] (n1) at (0,-1) { $\rv{O}_2$ }; \node[fac,label=right:{$p(\rv{N}_1 \rightarrow \rv{O}_2)$}] at (0,0) {} edge (n) edge (n1); 
\node[fac, label=below:\textcolor{red}{$\rv{O}_2=w_{l-1}$}, red, fill]   at (0, -2) {} edge (n1);

\end{tikzpicture} 

\\
\begin{tikzpicture} 
\node[ext](x) at (0,1) {$\rv{N}_1$};
\node[fac] at (0,0) { $\nt{X},\textcolor{red}{(i,j)}$ } edge (x); 
\end{tikzpicture} 
$\xlongrightarrow{\pi_6}$
\begin{tikzpicture} 
\node[ext] (n) at (0,1) { $\rv{N}_1$ }; 
\node[var] (n1) at (-1,-0.5) {$\rv{N}_2$}; 
\node[var] (n2) at (1,-0.5) {$\rv{N}_3$}; 
\node[fac,label=right:{$p(\rv{N}_1\rightarrow \rv{N}_2 \rv{N}_3)$}] at (0,0.25) {} edge (n) edge (n1) edge (n2); 
\node[fac] at (-1,-1.5) {$\nt{X_4},\textcolor{red}{(i, k)}$} edge (n1);
\node[fac] at (1,-1.5) {$\nt{X}_5,\textcolor{red}{(k, j)}$} edge (n2);
\end{tikzpicture}
\\
\end{tabular}
}
\caption{}
\end{subfigure}
\caption{FGG representations of (a) HMMs and (b) PCFGs. Examples come from \citet{DBLP:conf/nips/0001R20}.
}
\label{fig:fgg}
\end{figure*}

We conduct experiments in unsupervised PCFG parsing and HMM language modeling. 
For PCFG induction, we manage to use 20 times more hidden states than \citet{yang-etal-2021-pcfgs}, obtaining much better unsupervised parsing performance.  For HMM language modeling, we achieve lower perplexity and lower inference complexity than \citet{chiu2021low}.

\section{Background}
\subsection{Factor graph grammar}
Factor graphs are fixed-sized and thus incapable of modeling substructures that repeat a variable number of times.  \citet{DBLP:conf/nips/0001R20} propose factor graph grammars (FGGs) to overcome this limitation, which are expressive enough to subsume HMMs and PCFGs. The main purpose of introducing FGGs in this work is to facilitate more intuitive presentation of our method, and to enable generalization beyond HMMs and PCFGs.

\subsubsection{Basics}
 We display necessary notations and concepts of FGGs \cite[][\text{Def. 1,2,5,6,8}]{DBLP:conf/nips/0001R20}.
\paragraph{Definition 1.} A hypergraph is a tuple $\left(V, E, a t t, l a b^{V}, l a b^{E}\right)$ where 
\begin{itemize}
    \item $V$ and $E$ are finite set of nodes and hyperedges.
    \item $att: E \rightarrow V^{\star}$ maps each hyperedge to zero or more (not necessarily distinct) endpoint nodes.
    \item $lab^V: V \rightarrow L^V$  assigns labels to nodes.
    \item $lab^E: E \rightarrow L^E$ assigns labels to edges.
\end{itemize}

\paragraph{Definition 2.} A factor graph is a hypergraph with mappings $\Omega$ and $F$ where
\begin{itemize}
    \item  $\Omega$ maps node labels to sets of possible values. $\Omega(v) \triangleq \Omega(lab^V(v))$.
    \item  $F$ maps edge labels to functions. $F(e) \triangleq F(lab^E(e))$ is of type $\Omega(v_1) \times \cdots \times \Omega(v_k)$ where $att(e) = v_1 \cdots v_k$.
\end{itemize}
In the terminology of factor graphs, a node $v$ with its domain $\Omega(v)$ is a \textit{variable}, and an hyperedge $e$ with $F(e)$ is a \textit{factor}.  We typically use $\rv{T}, \rv{N}, \rv{O}$ to denote hidden state, nonterminal state and observation variables for HMMs and PCFGs.

\paragraph{Definition 3.} A hypergraph fragment is a tuple $(V, E, att, lab^{V}, lab^{E}, ext)$ where
\begin{itemize}
    \item $(V, E, att, lab^{V}, lab^{E})$ is a hypergraph.
    \item $ext \in V^{\star}$ is a set of zero or more external nodes and each of which can be seen as a connecting point of this hypergraph fragment with another fragment.
\end{itemize}
\paragraph{Definition 4.} A hyperedge replacement graph grammar (HRG) \cite{DBLP:conf/gg/DrewesKH97} is a tuple $(N, T, P, S)$ where 
\begin{itemize}
    \item $N, T \subset L^E$ is finite set of nonterminal and terminal symbols. $N \cap T = \emptyset$.
    \item $P$ is a finite set of rules ($X \rightarrow R$) where 
             $X \in N$ and $R$ is a hypergraph fragment with edge labels in $N \cup T$ \footnote{Note that, for the lhs of $P$, \citet{DBLP:conf/nips/0001R20} also draw their endpoint nodes using external node notations. We follow this practice.}.
    \item $S\in N$ is the start symbol.
\end{itemize}

\paragraph{Definition 5.}
A HRG with mapping $\Omega, F$ (Def. 2) is referred to as an FGG. In particular, $F$ is defined on terminal edge labels $T$ only. 

\paragraph{Notations.}
\begin{itemize}
    \item \begin{tikzpicture} \node[var] (t) at (0, 0) {$\rv{N}$}; \end{tikzpicture} : variable node.  \begin{tikzpicture} \node[ext] (t) at (0, 0) {$\rv{N}$}; \end{tikzpicture} : external node.
            \item \begin{tikzpicture}
            \node[var] (t2) at (-1, 0) {$ \quad$};
            \node[fac] (t) at (0, 0) {$\rv{X_e}$} edge(t2);
            \end{tikzpicture} :
             hyperedge $e$ with label $X\in N$.  \begin{tikzpicture} \node[var] (t) at (0, 0) {$\quad$}; \end{tikzpicture} indicates zero or more endpoint nodes.
            \item 
            \begin{tikzpicture} 
 \node (n) at (0,0.6) {}; 
\node (n1) at (-0.5,-0.5) {}; 
\node (n2) at (0.5,-0.5) {}; 
\node[fac,label=left:{$F(e)$}] at (0,0) {} edge (n) edge (n1) edge (n2); 
\end{tikzpicture}: factor $F(e)$.
\end{itemize}

Fig. \ref{fig:fgg} illustrates HGG representations of HMM and PCFG. 

\paragraph{Generative story.} An FGG starts with \begin{tikzpicture} \node[fac] (t) at (0, 0) {$\rv{S}$}; \end{tikzpicture}, repeatedly selects            \begin{tikzpicture}
            \node[] (t2) at (-0.75, 0) {};
            \node[fac] (t) at (0, 0) {$\rv{X_e}$} edge(t2);
            \end{tikzpicture}
and uses rule $X\rightarrow R$ from $P$ to replace $e$ with $R$, until no \begin{tikzpicture}
            \node[] (t2) at (-0.75, 0) {};
            \node[fac] (t) at (0, 0) {$\rv{X_e}$} edge(t2);
            \end{tikzpicture} exists.

\subsubsection{Conjunction}
The conjunction operation  \cite[][{Sec. 4}]{DBLP:conf/nips/0001R20} allows modularizing an FGG into two parts, one defining the model and the other defining a query. In this paper, we only consider querying the observed sentence  $w_0,\cdots,w_{n-1}$, 
which is exemplified by the red part of Fig. \ref{fig:fgg}. We sometimes omit the red part without further elaboration.

\subsubsection{Inference} 
  Denote $\xi$ as an assignment of all variables, $\Xi_{D}$ as the set of all assignments of factor graph $D$,  and $\mathcal{D}(G)$ as the set of all derivations of an FGG $G$, i.e., all factor graphs generated by $G$. 
an FGG $G$ assigns a score $w_{G}(D, \xi)$ to each $D \in \mathcal{D}(G)$  along with each $\xi \in \Xi_{D}$. A factor graph $D \in \mathcal{D}(G)$ assigns a score $w_{D}(\xi)$ to each $\xi \in \Xi_D$:
\begin{equation}
     w_D(\xi)= \prod_{e\in D} F(e)(\xi(v_1), \ldots, \xi(v_k))
\end{equation}
with $att(e) = v_1\cdots v_k$.  Notably, $w_{D}(\xi) \triangleq w_{G}(D, \xi)$. The inference problem is to compute the sum-product of $G$:
\begin{equation}
Z_{G} =\sum_{D \in \mathcal{D}(G)} \sum_{\xi \in \Xi_{D}} w_{G}(D, \xi) \label{eq:logZ} 
\end{equation}
To obtain $Z_G$, the key difficulty is in the marginalization over all derivations, since $\sum_{\xi \in \Xi_{D}} w_{D}(\xi)$ can be obtained by running standard variable elimination (VE) on factor graph $D$. To tackle this, \citet[][\text{Thm. 15}]{DBLP:conf/nips/0001R20} propose an extended VE.
For each $X \in N, \xi \in \Xi_{X}$ \footnote{$\Xi_{X}$ is defined as the set of assignments to the endpoints of an edge $e$ labeled X, so $\Xi_{X}=\Omega\left(\ell_{1}\right) \times \cdots \times \Omega\left(\ell_{k}\right)$ where $att(e) = v_1 \cdots v_k, lab^V(v_i)=\ell_i$.}, define $P^X$ as all rules in $P$ with left-hand side $X$, and then define:
\begin{equation}
  \psi_X(\xi) = \sum_{(X \rightarrow R) \in P^X} \tau_R(\xi).
  \label{eq:inference_1}
\end{equation}
for each rhs $R = (V, E_N \cup E_T, att, lab^V, lab^E, ext)$, where $E_N,E_T$ consist of nonterminal/terminal-labeled edges only, and  $\tau_R(\xi)$ is given by:
\begin{equation}
\begin{aligned}
\tau_{R}(\xi)=\sum_{\xi^{\prime} \in \Xi_{R} \atop \xi^{\prime}(e x t)=\xi} & \prod_{e \in E_{T}} F(e)\left(\xi^{\prime}(\operatorname{att}(e))\right) \\ &\prod_{e \in E_{N}} \psi_{l a b^{E}(e)}\left(\xi^{\prime}(\operatorname{att}(e))\right)
\label{eq:inference_2}
\end{aligned}
\end{equation}

This defines a recursive formula for computing $\psi_{S}$, i.e., $Z_G$.  Next, we will show how Eq. \ref{eq:inference_1}-\ref{eq:inference_2} recover the well-known inside algorithm.

\paragraph{Example: the inside algorithm.} 
Consider $\pi_6$ in Fig.\ref{fig:fgg}(b). All possible fragments $R$ (rhs of $\pi_6$) differs in the value of $k$, i.e., the splitting point, so we use $R_k$ to distinguish them. Then Eq. \ref{eq:inference_1} becomes:
\begin{equation}
 \psi_{X_{i,k}}(\xi) = \sum_{i<k<j} \tau_{R_k} (\xi)
 \label{eq:inference_3}
\end{equation}
Putting values into Eq. \ref{eq:inference_2}:
\begin{equation}
\tau_{R_k}(\xi) = \sum_{n_2, n_3} p(\xi, n_2, n_3) \psi_{X_{i,k}}(n_2) \psi_{X_{k, j}}(n_3)
\label{eq:inference_4}
\end{equation}
where $p$ denotes FGG rule probability $p(\rv{N}_1\rightarrow \rv{N}_2 \rv{N}_3)$. It is easy to see that $\psi_{X_{i,k}}$ is exactly the inside score of span $[i, k)$, and Eq. \ref{eq:inference_3}-\ref{eq:inference_4} recovers the recursive formula of the inside algorithm.

\paragraph{Remark.} Eq. \ref{eq:inference_2} can be viewed as unidirectional (from $e \in E_{N}$ to external nodes) belief propagation (BP) in the factor graph fragment $R$, where the incoming message is $\psi_{lab^E}(e)$ for $e \in E_{N}$, and the outcome of Eq. \ref{eq:inference_2} can be viewed as the message passed to the external nodes. The time complexity of message updates grows exponentially with the number of variables in the factors. Therefore, to decrease inference complexity, one may decompose large factors into smaller factors connected by new nodes, 
as shown in the next subsection.

\subsection{Tensor rank decomposition on factors}
Consider a factor $F(e)$ (Def. 2), it can be represented as an order-$k$ tensor in $\mathbb{R}^{N_1 \times \cdots \times N_{k}}$ where $N_{i} \triangleq |\Omega(v_i)|$. We can use tensor rank decomposition (aka. CPD) to decompose $F(e)$ into a weighted sum of outer products of vectors:
\[
F(e) =\sum_{q=1}^{r} \lambda_{q} \vect{w}_{e_1}^{q} \otimes \vect{w}_{e_2}^{q} \otimes \cdots \otimes \vect{w}_{e_k}^{q}
\]
where $r$ is the rank size; $\vect{w}_{e_k}^{q} \in \mathbb{R}^{N_k}$; $\otimes$ is outer product;  $\lambda_q$ is weight, which can be absorbed into $\{\vect{w}_{e_k}^{q}\}$ and we omit it throughout the paper.

\begin{figure}[t!]
    \centering
    \scalebox{0.75}{
\begin{tikzpicture} 
\node[var] (n1) at (0,1) { $\rv{v}_1$ }; 
\node[var] (n2) at (-1, -1) {$\rv{v_2}$};
\node[var] (n3) at (-0, -1) {$\rv{v_{..}}$};
\node[var] (n4) at (1, -1) {$\rv{v_{k}}$};
\node[fac, label=right:{$F(e)$}] at (0,0) {} edge (n2) edge (n3) edge(n4) edge(n1);
\end{tikzpicture}
$\rightarrow$
\begin{tikzpicture} 
\node[var] (n1) at (0,1) { $\rv{v}_1$ }; 
\node[var] (n2) at (-1, -2) {$\rv{v_2}$};
\node[var] (n3) at (-0, -2) {$\rv{v_{..}}$};
\node[var] (n4) at (1, -2) {$\rv{v_{k}}$};
\node[var] (r) at (0, -0.5) {$\rv{R}$}; 
\node[fac, label=right:{$F(e_1)$}] at ($(n1)!0.5!(r)$) {} edge(r) edge(n1);
\node[fac, label=left:{$F(e_2)$}] at ($(n2)!0.5!(r)$) {} edge(r) edge(n2);
\node[fac] at ($(n3)!0.5!(r)$) {} edge(r) edge(n3);
\node[fac, label=right:{$F(e_k)$}] at ($(n4)!0.5!(r)$) {} edge(r) edge(n4);
\end{tikzpicture}
}
    \caption{Using CPD to decompose a factor can be seen as adding a new node.}
    \label{fig:transform}
\end{figure}
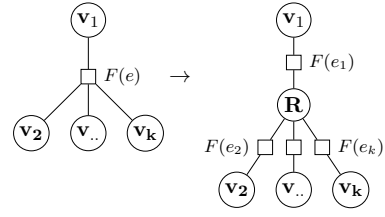

 \citet[][\text{Sec. 4.1}]{DBLP:journals/corr/abs-2010-09283} show that BP can be written in the following matrix form when applying CPD on factors:
\begin{align}
\vect{m}_{e i} &=\vect{W}_{e_i}^{T}\left(\odot_{j \in N(e) \backslash \mathrm{i}} \vect{W}_{e_j} \vect{n}_{j e}\right) 
\label{low-rank-message}
\\
\vect{n}_{i e} &=\odot_{c \in N(i) \backslash e} \vect{m}_{c i}
\end{align}
where $\vect{m}_{e i} \in \mathbb{R}^{N_i}$ is factor-to-node message; $\vect{n}_{i e} \in \mathbb{R}^{N_i}$ is node-to-factor message; $N(\cdot)$ indicates neighborhood ;  $\vect{W}_{e_j}=[\vect{w}_{e_j}^{1}, \cdots,  \vect{w}_{e_j}^{r}]^{T} \in \mathbb{R}^{r \times m}$; $\odot$ is  element-wise product. We remark that this amounts to replacing the large factor $F(e)$ with smaller factors $\{F(e_i)\}$ connected by a new node $\vect{R}$ that represents rank, where each $F(e_i)$ can be represented as $\vect{W}_{e_i}$. Fig. \ref{fig:transform} illustrates this intuition. We refer to $\vect{R}$ as rank nodes and others as state nodes thereafter.

\section{Low-rank structured inference}
\label{sec: low-rank}

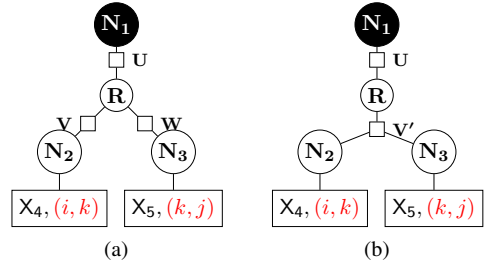
\begin{figure}[t!]
    \centering
    \scalebox{0.75}{
\begin{tikzpicture} 
\node (n1) at (0,-3) { \text{(a)} }; 
\node[ext] (n1) at (0,1) { $\rv{N_1}$ }; 
\node[var] (n2) at (-1, -1.25) {$\rv{N_2}$};
\node[var] (n4) at (1, -1.25) {$\rv{N_3}$};
\node[var] (r) at (0, -0.25) {$\rv{R}$}; 
\node[fac, label=right:{$\vect{U}$}] at ($(n1)!0.5!(r)$) {} edge(r) edge(n1);
\node[fac, label=left:{$\vect{V}$}] at ($(n2)!0.5!(r)$) {} edge(r) edge(n2);
\node[fac, label=right:{$\vect{W}$}] at ($(n4)!0.5!(r)$) {} edge(r) edge(n4);
\node[fac] (r) at (-1, -2.25) {$\nt{X_4},\textcolor{red}{(i,k)}$} edge(n2); 
\node[fac] (r) at (1, -2.25) {$\nt{X_5},\textcolor{red}{(k,j)}$} edge(n4); 
\end{tikzpicture}
 \qquad
\begin{tikzpicture} 
\node (n1) at (0,-3) { \text{(b)} }; 
\node[ext] (n1) at (0,1) { $\rv{N_1}$ }; 
\node[var] (n2) at (-1, -1.25) {$\rv{N_2}$};
\node[var] (n4) at (1, -1.25) {$\rv{N_3}$};
\node[var] (r) at (0, -0.25) {$\rv{R}$}; 
\node[fac, label=right:{$\vect{U}$}] at ($(n1)!0.5!(r)$) {} edge(r) edge(n1);
\node[fac, label=right:{$\vect{V^{\prime}}$}] at (0, -0.85) {} edge(r) edge(n4) edge(n2);
\node[fac] (r) at (-1, -2.25) {$\nt{X_4},\textcolor{red}{(i,k)}$} edge(n2); 
\node[fac] (r) at (1, -2.25) {$\nt{X_5},\textcolor{red}{(k,j)}$} edge(n4); 
\end{tikzpicture}
}
    \caption{Representations of the rhs of $\pi_6$ (Fig. \ref{fig:fgg}) after decomposition. (a): TD-PCFG \cite{cohen-etal-2013-approximate, yang-etal-2021-pcfgs}. (b): LPCFG \cite{chiu2021low}. }
    \label{fig:pcfgs}
\end{figure}
 In this section, we recover the accelerated inside algorithms of TD-PCFG \cite{cohen-etal-2013-approximate, yang-etal-2021-pcfgs} and LPCFG \cite{chiu2021low} in an intuitive and unifying manner using the FGG notations. The accelerated forward algorithm of LHMM \cite{chiu2021low} can be derived similarly.
 
Denote $\mathsf{T} \in \mathbb{R}^{m \times m \times m}$ as the tensor representation of $p(\rv{N}_1\rightarrow \rv{N}_2 \rv{N}_3)$ , and $\vect{\alpha}_{i, j} \in \mathbb{R}^{m}$ as the inside score of span $[i, j)$. 
\citet{cohen-etal-2013-approximate} and \citet{yang-etal-2021-pcfgs} use CPD to decompose $\mathsf{T}$, i.e., let $\mathsf{T} = \sum_{q=1}^{r} \vect{u}_{q} \otimes \vect{v}_{q} \otimes \vect{w}_{q}$ where $\vect{u}_{q}, \vect{v}_{q}, \vect{w}_{q} \in \mathbb{R}^{m}$. Denote $\vect{U}, \vect{V}, \vect{W} \in \mathbb{R}^{r \times m}$ as the resulting matrices of stacking all  $\vect{u}_{q}, \vect{v}_{q}, \vect{w}_{q}$, \citet{cohen-etal-2013-approximate} derived the recursive form:

\begin{align}
\vect{\alpha}_{i, j} &= \sum_{k = i+1}^{j - 1} \vect{U}^{T} \left(\left(\vect{V}{\vect{\alpha}_{i, k}}\right)\odot \left(\vect{W}{\vect{\alpha}_{k, j}}\right)\right)
\label{eq:pcfg_1}
\\
&= \vect{U}^{T} \sum_{k = i+1}^{j - 1}  \left(\left(\vect{V}{\vect{\alpha}_{i, k}}\right)\odot \left(\vect{W}{\vect{\alpha}_{k, j}}\right)\right)
\label{eq:pcfg_2}
\end{align}
Eq. \ref{eq:pcfg_1} can be derived automatically by combining Eq. \ref{low-rank-message} (or Fig. \ref{fig:pcfgs} (a)) and Eq. \ref{eq:inference_3}-\ref{eq:inference_4}.  \citet{cohen-etal-2013-approximate} note that $\vect{U^T}$ can be extracted to the front of the summation (Eq. \ref{eq:pcfg_2}), and $\vect{V}\vect{\alpha}_{i, k}, \vect{W}{\vect{\alpha}_{k, j}}$ can be cached and reused, leading to further complexity reduction. The resulting inside algorithm time complexity is $O(n^3r + n^2mr)$.

Recently, \citet{chiu2021low} use low-rank matrix decomposition to accelerate PCFG inference. They first perform  tensor matricization to flatten $\mathsf{T}$ to $\mathsf{T}^{\prime} \in \mathbb{R}^{m \times m^2}$ , and then let $\mathsf{T}^{\prime} = \vect{U}^{T} \vect{V}$ where $\vect{U} \in \mathbb{R}^{r \times m}, \vect{V} \in \mathbb{R}^{r \times m^2}$. By un-flattening $\vect{V}$ to $\vect{V^{\prime}} \in \mathbb{R}^{r \times m \times m}$, their accelerated inside algorithm has the following recursive form:
\begin{align}
\vect{\alpha}_{i, j} &=  \sum_{k = i+1}^{j - 1} \vect{U}^{T} \left(\vect{V^{\prime}}\cdot \vect{\alpha}_{k, j} \cdot
\vect{\alpha}_{i, k} \right)
\label{eq:pcfg_3}
\\
&= \vect{U}^{T} \sum_{k = i+1}^{j - 1}  \left(\vect{V}^{\prime} \cdot \vect{\alpha}_{k, j} \cdot \vect{\alpha}_{i, k} \right)
\label{eq:pcfg_4}
\end{align}
Eq. \ref{eq:pcfg_3} can be derived by combining Fig. \ref{fig:pcfgs} (b) and Eq. \ref{eq:inference_3}-\ref{eq:inference_4}. The resulting inside time complexity is $O(n^3m^2r+n^2mr)$, which is higher than that of TD-PCFG.

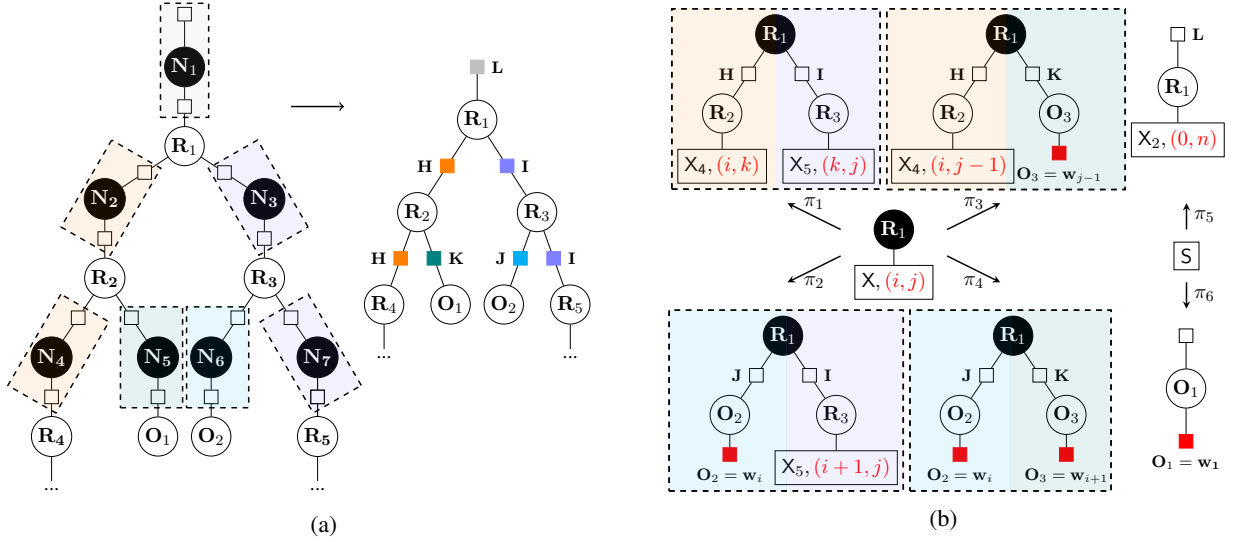
\begin{figure*}[tb!]
	\begin{subfigure}{0.53\linewidth}
	\scalebox{0.7}{
\begin{tikzpicture} 
\node[ext] (n1) at (0,1) { $\rv{N}_1$ }; 
\node[var] (r1) at (0,-0.5) {$\rv{R}_1$}; 
\node[fac,label=right:{}] (f0) at (0,2) {} edge (n1);
\node[fac] (f-1) at (0,0.25) {} edge (n1) edge (r1);
\node[ext] (n2) at (-1.5, -1.5) {$\rv{N_2}$};
\node[ext] (n3) at (1.5, -1.5) {$\rv{N_3}$};
\node[fac] (f1) at ($(r1)!0.5!(n2)$) {} edge (r1) edge (n2);
\node[fac]  (f3) at ($(r1)!0.5!(n3)$) {} edge (r1) edge (n3);
\node[var] (r2) at (-1.5, -3) {$\rv{R_2}$}; 
\node[fac] (f2) at ($(r2)!0.5!(n2)$) {} edge (r2) edge (n2);
\node[ext] (n4) at (-2.5, -4.5) {$\rv{N_4}$};
\node[ext] (n6) at (0.5, -4.5) {$\rv{N_6}$};
\node[ext] (n7) at (2.5, -4.5) {$\rv{N_7}$};
\node[ext] (n5) at (-0.5, -4.5) {$\rv{N_5}$};
\node[var] (w3) at (-0.5, -6) {$\rv{O}_1$};
\node[var] (w4) at (0.5, -6) {$\rv{O}_2$};
\node[var] (r3) at (1.5, -3) {$\rv{R_3}$}; 
\node[var] (r4) at (-2.5, -6) {$\rv{R_4}$}; 
\node[var] (r5) at (2.5, -6) {$\rv{R_5}$}; 

\node[var] (n11) at (5.5,0) { $\rv{R}_1$ }; 
\node[var] (n12) at (4.375,-1.75) { $\rv{R}_2$ }; 
\node[var] (n13) at (6.625, -1.75) { $\rv{R}_3$ }; 
\node[var] (n14) at (3.75,-3.5) { $\rv{R}_4$ }; 
\node[var] (n15) at (5,-3.5) { $\rv{O}_1$ }; 
\node[var] (n16) at (6,-3.5) { $\rv{O}_2$ };  
\node[var] (n17) at (7.25,-3.5) { $\rv{R}_5$ }; 
\node[] (q3) at (3.75, -4.5) {...};
\node[] (q4)at (7.25, -4.5) {...};
\draw[] (q3) -- (n14);
\draw[] (q4) -- (n17);
\node[fac, fill, lightgray, label=right:{$\vect{L}$}] () at (5.5, 1) {} edge (n11);
\node[fac, fill, orange, label=left:{$\vect{H}$}] (e1) at ($(n11)!0.5!(n12)$) {} edge (n11) edge (n12);
\node[fac, fill, blue!50, label=right:{$\vect{I}$}] (e2) at ($(n11)!0.5!(n13)$) {} edge (n11) edge (n13);
\node[fac, fill, orange,label=left:{$\vect{H}$}] (e3) at ($(n12)!0.5!(n14)$) {} edge (n12) edge (n14);
\node[fac, fill, teal, label=right:{$\vect{K}$}] (e4) at ($(n12)!0.5!(n15)$) {} edge (n12) edge (n15);
\node[fac, fill, cyan, label=left:{$\vect{J}$}] (e5) at ($(n13)!0.5!(n16)$) {} edge (n13) edge (n16);
\node[fac, fill, blue!50, label=right:{$\vect{I}$}] (e6) at ($(n13)!0.5!(n17)$) {} edge (n13) edge (n17);

\node[fac] (e7) at ($(n7)!0.5!(r5)$) {} edge (n7) edge (r5);

\node[fac] (f5) at ($(r2)!0.5!(n4)$) {} edge (r2) edge (n4);
\node[fac] (f10)at ($(r2)!0.5!(n5)$) {} edge (r2) edge (n5);
\node[fac] (f4) at ($(n3)!0.5!(r3)$) {} edge (n3) edge (r3);
\node[fac] (f12) at ($(r3)!0.5!(n6)$) {} edge (r3) edge (n6);
\node[fac] (f7) at ($(r3)!0.5!(n7)$) {} edge (r3) edge (n7);
\node[fac] (f9) at ($(n5)!0.5!(w3)$) {} edge (n5) edge (w3);
\node[fac] (f11) at ($(n6)!0.5!(w4)$) {} edge (n6) edge (w4);
\node[fac] (f6) at ($(n4)!0.5!(r4)$) {} edge (n4) edge (r4);
\node[fac] (f8) at ($(n7)!0.5!(r5)$) {} edge (n7) edge (r5);
\node[] (q1) at (-2.5, -7) {...};
\node[] (q2)at (2.5, -7) {...};

\node [draw, fill=orange,  fill opacity=0.1,  thick, dashed, rotate fit=60, fit=(f1)(n2)(f2),  inner sep = 3pt] {};    
\node [draw, fill=orange, fill opacity=0.1, thick, dashed, rotate fit=60, fit=(f5)(n4)(f6),  inner sep = 3pt] {};    
\node [draw, fill=blue!50,fill opacity=0.1,  thick, dashed, rotate fit=-60, fit=(f3)(n3)(f4),  inner sep = 3pt] {};    
\node [draw, fill=blue!50,fill opacity=0.1,  thick, dashed, rotate fit=-60, fit=(f7)(n7)(f8),  inner sep = 3pt] {};    
\node [draw, fill=teal,fill opacity=0.1,  thick, dashed, fit=(f9)(n5)(f10),  inner sep = 2pt] {};    
\node [draw, fill=cyan,fill opacity=0.1,  thick, dashed, fit=(f11)(n6)(f12),  inner sep = 2pt] {};    
\node [draw, fill=lightgray, fill opacity=0.1, thick, dashed, fit=(f-1)(n1)(f0),  inner sep = 2pt] {};    

\draw[->, thick] (2,0.25) -- (3,0.25);
\draw[] (r4) -- (q1);
\draw[] (r5) -- (q2);


\end{tikzpicture}
}
\caption{}
\end{subfigure}
\begin{subfigure}{0.48\linewidth}
\scalebox{0.7}{
\begin{tabular}{c}
\begin{tikzpicture} 
[fill left half/.style={path picture={\fill[#1] (path picture bounding box.south west)
  rectangle (path picture bounding box.north);}},
fill right half/.style={path picture={\fill[#1] (path picture bounding box.south east)
  rectangle (path picture bounding box.north);}}]

\node[ext] (n) at (0,1) { $\rv{R}_1$ }; 
\node[var] (n1) at (-1,-0.5) {$\rv{R}_2$}; 
\node[var] (n2) at (1,-0.5) {$\rv{R}_3$}; 
\node[fac, label=left:{$\vect{H}$}] at ($(n)!0.5!(n1)$) {} edge (n) edge (n1);
\node[fac,label=right:{$\vect{I}$}] at ($(n)!0.5!(n2)$) {} edge (n) edge (n2);
\node[fac] (f1) at (-1,-1.5) {$\nt{X_4},\textcolor{red}{(i, k)}$} edge (n1);
\node[fac] (f2) at (1,-1.5) {$\nt{X}_5,\textcolor{red}{(k, j)}$} edge (n2);
\node [draw, fill left half=orange, fill opacity=0.1,  thick, dashed,  fit=(f1)(f2)(n),  inner sep = 3pt] {};    
\node [draw, fill right half=blue!50, fill opacity=0.1,  thick, dashed,  fit=(f1)(f2)(n),  inner sep = 3pt] {};    
\end{tikzpicture}
\begin{tikzpicture} 
[fill left half/.style={path picture={\fill[#1] (path picture bounding box.south west)
  rectangle (path picture bounding box.north);}},
fill right half/.style={path picture={\fill[#1] (path picture bounding box.south east)
  rectangle (path picture bounding box.north);}}]
\node[ext] (n) at (0,1) { $\rv{R}_1$ }; 
\node[var] (n1) at (-1,-0.5) {$\rv{R}_2$}; 
\node[var] (n2) at (1,-0.5) {$\rv{O}_3$}; 
\node[fac,label=left:{$\vect{H}$}]  at ($(n)!0.5!(n1)$) {} edge (n) edge (n1);
\node[fac,label=right:{$\vect{K}$}] at ($(n)!0.5!(n2)$) {} edge (n) edge (n2);
\node[fac] at (-1,-1.5) {$\nt{X_4},\textcolor{red}{(i, j-1)}$} edge (n1);
\node[fac, red, fill, label=below:{$\rv{O}_3 = \vect{w}_{j-1}$}] at (1,-1.25) {} edge (n2);
\node[] (f1) at (-2,-1.7) {};
\node[] (f2) at (2,-1.7) {};
\node [draw, fill left half=orange, fill opacity=0.1,  thick, dashed,  fit=(f1)(f2)(n),  inner sep = 3pt] {};    
\node [draw, fill right half=teal, fill opacity=0.1,  thick, dashed,  fit=(f1)(f2)(n),  inner sep = 3pt] {};    
\end{tikzpicture}
\begin{tikzpicture} 
\node[var] (n) at (0,0) { $\rv{R}_1$ }; 
\node[fac,label=right:{$\vect{L}$}] at (0,1) {} edge (n);
\node[fac] at (0,-1) {$\nt{X}_2, \textcolor{red}{(0, n)}$} edge (n); 
\end{tikzpicture} 
\\
    \begin{tikzpicture} 
\node[] (x) at (-5,0) {};
\node[] (x) at (5,0) {};
\node[ext](x) at (-1,1) {$\rv{R}_1$};
\node[fac] at (-1,0) { $\nt{X},\textcolor{red}{(i,j)}$ } edge (x); 
\node[fac] at (4.5,0.5) { $\nt{S}$ }; 
\draw [arrow] (-2, 1)-- node[above] {$\pi_1$}  (-3, 1.5) ;
\draw [arrow] (-2, 0.5)-- node[below] {$\pi_2$}  (-3, 0) ;
\draw [arrow] (0, 1)-- node[above] {$\pi_3$}  (1, 1.5) ;
\draw [arrow] (0, 0.5)-- node[below] {$\pi_4$}  (1, 0) ;
\draw [arrow] (4.5, 1) -- node[right] {$\pi_5$} (4.5, 1.5);
\draw [arrow] (4.5, 0) -- node[right] {$\pi_6$} (4.5, -0.5);
\end{tikzpicture} 
\\
\begin{tikzpicture} 
[fill left half/.style={path picture={\fill[#1] (path picture bounding box.south west)
  rectangle (path picture bounding box.north);}},
fill right half/.style={path picture={\fill[#1] (path picture bounding box.south east)
  rectangle (path picture bounding box.north);}}]
\node[ext] (n) at (0,1) { $\rv{R}_1$ }; 
\node[var] (n1) at (-1,-0.5) {$\rv{O}_2$}; 
\node[var] (n2) at (1,-0.5) {$\rv{R}_3$}; 
\node[fac,label=left:{$\vect{J}$}] at ($(n)!0.5!(n1)$) {} edge (n) edge (n1);
\node[fac,label=right:{$\vect{I}$}] at ($(n)!0.5!(n2)$) {} edge (n) edge (n2);
\node[fac, red, fill, label=below:{$\rv{O}_2 = \vect{w}_{i}$}]  at (-1,-1.25) {} edge (n1);
\node[fac](f2) at (1,-1.5) {$\nt{X}_5,\textcolor{red}{(i+1, j)}$} edge (n2);
\node[] (f1) at (-1.9,-1.7) {};
\node [draw, fill left half=cyan, fill opacity=0.1,  thick, dashed,  fit=(f1)(f2)(n),  inner sep = 3pt] {};    
\node [draw, fill right half=blue!50, fill opacity=0.1,  thick, dashed,  fit=(f1)(f2)(n),  inner sep = 3pt] {};    
\end{tikzpicture}
\begin{tikzpicture} 
[fill left half/.style={path picture={\fill[#1] (path picture bounding box.south west)
  rectangle (path picture bounding box.north);}},
fill right half/.style={path picture={\fill[#1] (path picture bounding box.south east)
  rectangle (path picture bounding box.north);}}]
\node[ext] (n) at (0,1) { $\rv{R}_1$ }; 
\node[var] (n1) at (-1,-0.5) {$\rv{O}_2$}; 
\node[var] (n2) at (1,-0.5) {$\rv{O}_3$}; 
\node[fac,label=left:{$\vect{J}$}] at ($(n)!0.5!(n1)$) {} edge (n) edge (n1);
\node[fac,label=right:{$\vect{K}$}] at ($(n)!0.5!(n2)$) {} edge (n) edge (n2);
\node[fac, red, fill, label=below:{$\rv{O}_2 = \vect{w}_{i}$}] at (-1,-1.25) {} edge (n1);
\node[fac, red, fill, label=below:{$\rv{O}_3 = \vect{w}_{i+1}$}] at (1,-1.25) {} edge (n2);
\node[] (f1) at (-1.7,-1.7) {}; 
\node[] (f2) at (1.6,-1.7) {}; 
\node[var] (n3) at (3.25,0) { $\rv{O}_1$ }; 
\node[fac,label=right:{}] at (3.25,1) {} edge (n3);
\node[fac, red, fill, label=below:{$\rv{O}_1 = \vect{w_1}$}] at (3.25,-1){} edge (n3); 
\node [draw, fill left half=cyan, fill opacity=0.1,  thick, dashed,  fit=(f1)(f2)(n),  inner sep = 3pt] {};    
\node [draw, fill right half=teal, fill opacity=0.1,  thick, dashed,  fit=(f1)(f2)(n),  inner sep = 3pt] {};    
\end{tikzpicture}
\end{tabular}
}
\caption{}
\end{subfigure}

\caption{(a): illustration of marginalizing state nodes $\rv{N}$.  (b): rule set of the new FGG. $\pi_1$ can be applied when $k\ne i+1$ and $k+1\ne j$; $\pi_2$ and $\pi_3$ can be applied when $i\ne j-1$; $\pi_4$ can be applied when $j=i+2$.
}
\label{fig:new_pcfg_rule}
\end{figure*}

 When learning a PCFG and a HMM, there is no need to first learn $\mathsf{T}$ and then perform decomposition on $\mathsf{T}$. Instead, one can learn the decomposed matrices (e.g., $\vect{U}, \vect{V}$) to learn $\mathsf{T}$ implicitly. During inference, one can follow Eq. \ref{eq:pcfg_2} or \ref{eq:pcfg_4} without the need to reconstruct $\mathsf{T}$.

\paragraph{Validity of probability.}  The remaining problem is to ensure that $\mathsf{T}$ is a valid probability tensor (i.e., being nonnegative and properly normalized) when learning it implicitly.  \citet{yang-etal-2021-pcfgs} essentially transform Fig. \ref{fig:pcfgs}(a) into a Bayesian network, adding directed arrows $\rv{N_1} \rightarrow \rv{R}, \rv{R} \rightarrow \rv{N_2}, \rv{R} \rightarrow \rv{N_3}$. This is equivalent to requiring that $\vect{V}, \vect{W}$ are nonnegative and column-wise normalized and $\vect{U}$ is nonnegative and row-wise normalized, as described in \citet[][\text{Thm.~1}]{yang-etal-2021-pcfgs}. One can apply the Softmax re-parameterization to enforce such requirement, which is more convenient in end-to-end learning. \citet{chiu2021low} replace the local normalization of \citet{yang-etal-2021-pcfgs} with global normalization, and we refer readers to their paper for more details. We adopt the strategy of \citet{yang-etal-2021-pcfgs} in this work.

\section{Rank-space modeling and inference}
\subsection{Rank-space inference with B-FGGs}
Interestingly, when applying CPD on factors and if the rank size is smaller than the state size, we can even obtain better inference time complexities for a subset of FGGs which we refer to as B-FGGs. 

We call a hyperedge a B-edge if its head contains exactly one node.
B-graphs \cite{DBLP:journals/dam/GalloLP93} are a subset of directed hypergraphs whose hyperedges are all B-edges. Many dynamic programming algorithms can be formulated through B-graphs \cite{DBLP:conf/iwpt/KleinM01,huang-2008-advanced, DBLP:journals/corr/AzumaSM17, chiu2021low,  fu2021scaling}, including the inference algorithms of many structured models, e.g., HMMs, Hidden Semi-Markov Models (HSMMs), and PCFGs. We follow the concept of B-graphs to define B-FGGs.

\paragraph{Definition 6.} A hypergraph fragment is a B-hypergraph fragment iff. there is exactly one external node and there is no nonterminal-labeled hyperedge connecting to it. An FGG is a B-FGG iff. all rhs of its rules are B-hypergraph fragments.

\smallskip
It is easy to see that the aforementioned models are subsumed by B-FGGs. We can design a family of accelerated inference algorithms for B-FGGs based on the following strategy. (1) If there are multiple factors within a hypergraph fragment, merge them into a single factor. Then apply CPD on the single factor, thereby introducing rank nodes. (2) Find repeated substructures that take rank nodes as external nodes.  Marginalize all state nodes to derive new rules. (3) Design new inference algorithms that can be carried out in the rank space based on the general-purpose FGG inference algorithm and the derived new rules. We give two examples, the rank-space inside algorithm and the rank-space forward algorithm, in the following two subsections to help readers understand this strategy.

\subsection{The rank-space inside algorithm}
\label{sec:rank-inside}
Consider an B-FGG $G$ shown in Fig.\ref{fig:fgg}(b) and replace the rhs of $\pi_6$ with Fig. \ref{fig:pcfgs}(a), i.e., we use CPD to decompose binary rule probability tensor. Besides $\vect{U},\vect{V},\vect{W} \in \mathbb{R}^{r \times m}$ defined in Sec. \ref{sec: low-rank}, we define the start rule probability vector as $\vect{s}\in \mathbb{R}^{m \times 1}$, and the unary rule probability matrix as $\vect{E} \in \mathbb{R}^{o \times m}$ where $o$ is the vocabulary size.

Fig. \ref{fig:new_pcfg_rule}(a) is an example (partial) factor graph $D$ generated by $G$. We highlight substructures of interest with dashed rectangles. Each substructure consists of a node $\rv{N}$ and two factors connecting to it. $\rv{N}$ is an external node connecting two hypergraph fragments which contain the two factors respectively. For each substructure, we can marginalize the state 
node $\rv{N}$ out, merging the two factors into a single one. After marginalizing all state 
nodes, we obtain a (partial) factor graph $D^{\prime}$ shown in the right of Fig. \ref{fig:new_pcfg_rule}(a) where $\vect{H} = \vect{V}\vect{U}^{T}, \vect{I} = \vect{W}\vect{U}^{T}, \vect{J} = \vect{V}\vect{E}^{T}, \vect{K}=\vect{W}\vect{E}^{T}$%
, $\vect{L}=(\vect{U}\vect{s})^{T}$. We denote this transformation as $\mathcal{M}(D)=D^{\prime}$. It is worth mentioning that $\vect{H}, \vect{I}, \vect{J}, \vect{K}, \vect{L}$ are computed only \emph{once} and then reused multiple times during inference, which is the key to reduce the time complexity. 

Then we define a new B-FGG $G^{\prime}$ with rules shown in Fig \ref{fig:new_pcfg_rule}(b). It is easy to verify that for each $D \in \mathcal{D}(G)$, we have $\mathcal{M}(D) \in \mathcal{D}(G^{\prime})$, and vice versa.  Moreover, we have: \[ \sum_{\xi \in \Xi_{D}} w_{G}(D, \xi) = \sum_{\xi \in \Xi_{\mathcal{M}(D)}} w_{G^{\prime}}(\mathcal{M}(D), \xi)\] 
because marginalizing hidden variables does not affect the result of sum-product inference. Therefore,  $Z_G = Z_{G^{\prime}}$ (Eq. \ref{eq:logZ}).

We can easily derive the inference (inside) algorithm of $G^{\prime}$ by following Eq. \ref{eq:inference_1}-\ref{eq:inference_2} and Fig. \ref{fig:new_pcfg_rule}(b) \footnote{$\pi_6$ is used for generating sentences of length 1, we do not consider this in the following derivation of the inside algorithm to reduce clutter.}. Let $\vect{\alpha}_{i, j}\in \mathbb{R}^{r}$ denote the rank-space inside score for span $[i,j)$. When $j>i+2$:
\begin{align*}
\vect{\alpha}_{i, j} &= \overbrace{\sum_{i+1<k<j-1} (\vect{H} \vect{\alpha}_{i, k} \odot \vect{I}\vect{\alpha}_{k, j})}^{\text{from $\pi_{1}$ of Fig. \ref{fig:new_pcfg_rule}(b)}}  \\
                    & + \underbrace{\vect{J}_{:,w_{i}} \odot \vect{I} \vect{\alpha}_{i+1, j}}_{\text{from $\pi_{2}$}} + \underbrace{\vect{H}\vect{\alpha}_{i, j-1} \odot \vect{K}_{:,w_{j-1}}}_{\text{from $\pi_{3}$}}
\end{align*}
and when $j = i+2$, $\vect{\alpha}_{i, j} = \vect{J}_{:,w_{i}} \odot \vect{K}_{:,w_{i+1}}$ (from $\pi_{4})$. $w_j$ is the index of the $j$-th word of the input sentence in the vocabulary; $\vect{A}_{:,j}$ indicates the $j$-th column of $\vect{A}$.

\begin{figure*}[tb!]
\begin{subfigure}{.35\linewidth}
        \centering
        \scalebox{0.6}{
    \begin{tikzpicture}
\node[ext] (x) at (-2,0) {$\rv{T}_1$}; 
\node[var] (t) at (0,0) {$\rv{T}_2$}; 
\node[var] (w) at (0,-2) {$\rv{O}_3$};
\node[fac,label=above:{$p(\rv{T}_2\mid \rv{T}_1)$}] at (-1,0) {} edge (x) edge (t); 
\node[fac,label=left:{$p(\rv{O}_3 \mid \rv{T}_2)$}] at (0,-1) {} edge (t) edge (w);
\node[fac] at (1,0) {$\nt{X}_4$ } edge (t); 
\end{tikzpicture} 
    \begin{tikzpicture}
    \draw[arrow] (-1, -1) -- (0, -1);
\end{tikzpicture}
    \begin{tikzpicture}
\node[ext] (x) at (-1,0) {$\rv{T}_1$};
\node[var] (r) at (0.5,-0.75) {$\rv{R}_2$}; 
\node[var] (t) at (2,0) {$\rv{T}_3$}; 
\node[var] (w) at (0.5,-2.25) {$\rv{O}_4$};
\node[fac,label=above:{$\vect{U}$}] at ($(x)!0.5!(r)$) {} edge (x) edge (r); 
\node[fac, label=above:{$\vect{V}$}] at ($(r)!0.5!(t)$) {} edge (t) edge (r); 
\node[fac, label=right:{$\vect{W}$}] at ($(r)!0.5!(w)$) {} edge (w) edge (r); 
\node[fac] at (3,0) {$\nt{X}_4$ } edge (t); 
\end{tikzpicture} 
}
\caption{}
\end{subfigure}
\begin{subfigure}{.45\linewidth}
\centering
\scalebox{0.6}{
    \begin{tikzpicture}
\node[ext] (x) at (-1,0) {$\rv{T}_1$};
\node[fac, label=left:{$\rv{T_1}=\text{bos}$}] (f1) at (-1.75, 0) {} edge (x);
\node[var] (r) at (-0.25,-1.5) {$\rv{R}_2$}; 
\node[ext] (t) at (0.5,0) {$\rv{T}_4$}; 
\node[var] (w) at (-0.25,-3) {$\rv{O}_3$};

\node[var] (r2) at (1.25,-1.5) {$\rv{R}_5$}; 
\node[ext] (t2) at (2,0) {$\rv{T}_7$}; 
\node[var] (w2) at (1.25,-3) {$\rv{O}_6$};

\node[var] (r3) at (2.75,-1.5) {$\rv{R}_8$}; 
\node[ext] (t3) at (3.5,0) {$\rv{T}_{10}$}; 
\node[var] (w3) at (2.75,-3) {$\rv{O}_{9}$};
\node[fac,label=right:{$\rv{T_3}=\text{eos}$}] (f8) at (4.25, 0) {} edge (t3);

\node[fac] (f2) at ($(x)!0.5!(r)$) {} edge (x) edge (r); 
\node[fac] (f3) at ($(r)!0.5!(t)$) {} edge (t) edge (r); 
\node[fac] at ($(r)!0.5!(w)$) {} edge (w) edge (r); 

\node[fac] (f4) at ($(t)!0.5!(r2)$) {} edge (t) edge (r2); 
\node[fac] (f5) at ($(r2)!0.5!(t2)$) {} edge (t2) edge (r2); 
\node[fac] at ($(r2)!0.5!(w2)$) {} edge (w2) edge (r2); 

\node[fac](f6) at ($(t2)!0.5!(r3)$) {} edge (t2) edge (r3); 
\node[fac] (f7) at ($(r3)!0.5!(t3)$) {} edge (t3) edge (r3); 
\node[fac] at ($(r3)!0.5!(w3)$) {} edge (w3) edge (r3); 


\node [draw, fill=orange,  fill opacity=0.1,  thick, dashed,  fit=(f1)(x)(f2),  inner sep = 3pt] {};    
\node [draw, fill=blue!50,  fill opacity=0.1,  thick, dashed,  fit=(f3)(t)(f4),  inner sep = 3pt] {};    
\node [draw, fill=blue!50,  fill opacity=0.1,  thick, dashed,  fit=(f5)(t2)(f6),  inner sep = 3pt] {};    
\node [draw, fill=cyan,  fill opacity=0.1,  thick, dashed,  fit=(f7)(t3)(f8),  inner sep = 3pt] {};    
\end{tikzpicture}
}
\caption{}
\end{subfigure}
\begin{subfigure}{.18\linewidth}
\centering
\scalebox{0.6}{
\begin{tikzpicture}
\node[] at (5, 1) {};
\node[var] (r4) at (5,0) {$\rv{R}_2$}; 
\node[var] (r5) at (6.5,0) {$\rv{R}_5$}; 
\node[var] (r6) at (8,0) {$\rv{R}_8$}; 

\node[var] (w-1) at (5,-1.5) {$\rv{O}_3$}; 
\node[var] (w-2) at (6.5,-1.5) {$\rv{O}_6$}; 
\node[var] (w-3) at (8,-1.5) {$\rv{O}_9$}; 

\node[fac] at ($(r6)!0.5!(w-3)$) {} edge (w-3) edge (r6); 
\node[fac] at ($(r5)!0.5!(w-2)$) {} edge (w-2) edge (r5); 
\node[fac] at ($(r4)!0.5!(w-1)$) {} edge (w-1) edge (r4); 
\node[fac] at ($(r4)!0.5!(w-1)$) {} edge (w-1) edge (r4);

\node[fac, fill=orange] at (4.25, -0) {} edge (r4); 
\node[fac, fill=cyan] at (8.75, 0) {} edge (r6); 

\node[fac, fill=blue!50]  at ($(r4)!0.5!(r5)$) {} edge (r4) edge (r5); 
\node[fac, fill=blue!50]  at ($(r5)!0.5!(r6)$) {} edge (r5) edge (r6); 

\end{tikzpicture} 
}
\caption{}
\end{subfigure}
    \caption{(a): merge the two factors into a single one, and apply CPD on the resulting factor. (b): factor graph of a HMM for sentences of length 3. (c): the resulting factor graph after marginalizing the state nodes.}
    \label{fig:new_hmm}
\end{figure*}
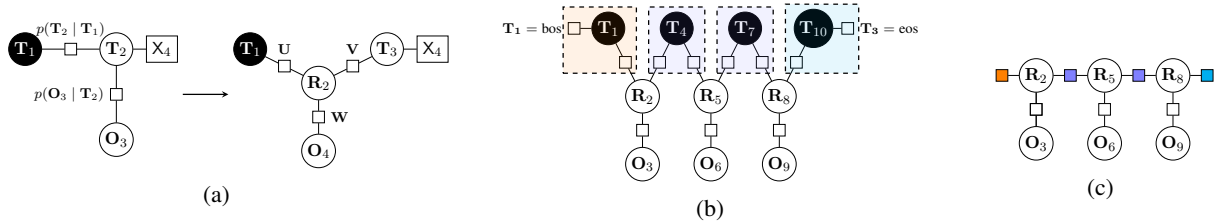
We note that, similar to \citet{cohen-etal-2013-approximate}, we can cache $\vect{H}\vect{\alpha}_{i, k}, \vect{I}\vect{\alpha}_{k,j}$ and reuse them to further accelerate inference \footnote{In fact, this is a typical application of the unfold-refold transformation \cite{eisner-blatz-2007, vieira-etal-2021-searching-efficient}.}. Denote $\vect{\alpha}_{i, j}^{L}, \vect{\alpha}_{i, j}^{R} \in \mathbb{R}^{r}$ as the inside scores of span $[i, j)$ serving as a left/right child of a larger span. Then we have:
\begin{align*}
\vect{\alpha}_{i, i+1}^{L} &= \vect{K}_{:, i} \qquad \vect{\alpha}_{i, i+1}^{R} = \vect{J}_{:, i}\\
\vect{\alpha}_{i, j}^{L} &= \vect{H} \vect{\alpha}_{i, j} \qquad \alpha_{i, j}^{R} = \vect{I} \vect{\alpha}_{i, j} \\
\vect{\alpha}_{i, j} &=  \sum_{i<k<j} (\vect{\alpha}_{i, k}^{L} \odot \vect{\alpha}_{k, j}^{R})
\end{align*}
and finally, $Z_{G^{\prime}}= \vect{L}\vect{\alpha}_{0, n}$.  We minimize $-\log Z_{G^{\prime}}$ using mini-batch gradient descent for unsupervised learning. The resulting inference complexity is $O(n^3r+n^2r^2)$\footnote{This does not take into account the one-time cost of computing $ \vect{H}, \vect{I}, \vect{J}, \vect{K}$ before inference.
}, which is lower than $O(n^3r+n^2mr)$ of TD-PCFG when $r < m$, enabling the use of a large state space for PCFGs in the low-rank setting.  

The key difference between the rank-space inference and the original state-space inference is that they follow different variable elimination orders. The former marginalizes all state nodes before performing inference and marginalizes rank nodes from bottom up during inference; whereas the later marginalizes both state and rank nodes alternately from bottom up during inference.

\paragraph{Parsing.} Low-rank inference does not support the Viterbi semiring\footnote{The Viterbi semiring is also known as the max-product semiring. \citet[][\text{Appd. C}]{chiu2021low} and \citet[][\text{Sec. 6}]{yang-etal-2021-pcfgs} have discussed this issue.}, inhibiting the use of CYK decoding. Therefore, we follow \citet{yang-etal-2021-pcfgs} to use Minimum Bayes-Risk decoding \citep{goodman-1996-parsing}. Specifically, we estimate the span marginals using auto-differentiation \cite{eisner-2016-inside,rush-2020-torch}, which has the same complexity as the inside algorithm. Then we use the CYK algorithm to find the final parse with the maximum number of expected spans in $O(n^3)$ time, similar to \citet{smith-eisner-2006-minimum}.

\paragraph{Implementation.} The implementation of the inside algorithm greatly influences the actual running speed. First, $O(n^2)$ out of $O(n^3)$ can be computed in parallel using parallel parsing techniques \cite{yi-etal-2011-efficient, canny-etal-2013-multi, DBLP:conf/ijcai/ZhangZL20, rush-2020-torch}. In this work, we adapt the efficient implementation of \citet{DBLP:conf/ijcai/ZhangZL20} for fast inside computation. Second, we adopt the \texttt{log-einsum-exp} trick \cite{DBLP:conf/icml/PeharzLVS00BKG20} to avoid expensive \texttt{log-sum-exp} operations on high-dimensional vectors, which reduces both GPU memory usage and total running time.

\subsection{The rank-space forward algorithm}
\label{sec:rank-forward}

Consider an B-FGG $G$ shown in Fig. \ref{fig:fgg} (a). We replace the rhs of $\pi_2$ by the hypergraph fragment in the right of Fig. \ref{fig:new_hmm}(a), i.e., we merge the factor $p(\rv{T}_2\mid \rv{T}_1)$ and $p(\rv{O}_3 \mid \rv{T}_2)$ into a single factor, which can be represented as $\mathsf{T}\in \mathbb{R}^{m \times m \times o}$ and can be decomposed into three matrices $\vect{U, V} \in \mathbb{R}^{r \times m}, \vect{W} \in \mathbb{R}^{r \times o}$ via CPD, where $m/o/r$ is the  state/vocabulary/rank size. Fig. \ref{fig:new_hmm}(b) gives an example factor graph of HMMs with sentences of length 3. Similar to previous subsection, we marginalize state nodes $\rv{T}$ to construct a new B-FGG $G^{\prime}$. The rule set of $G^{\prime}$ can be obtained by replacing all variable nodes $\rv{T}$ with $\rv{R}$ and modifying all factors accordingly, as one can 
easily infer from Fig. \ref{fig:new_hmm}(c). Inference with $G^{\prime}$ simply coincides with the forward algorithm, which has a $O(nr^2)$ time complexity and is lower than $O(nmr)$ of LHMM \cite{chiu2021low} when $r < m$.

\subsection{Neural parameterization}
We use neural networks to produce probabilities for all factors, which has been shown to benefit learning and unsupervised induction of syntactic structures \cite{jiang-etal-2016-unsupervised, he-etal-2018-unsupervised, kim-etal-2019-compound, han-etal-2019-enhancing,  jin-etal-2019-unsupervised, zhu-etal-2020-return, yang-etal-2020-second, yang-etal-2021-pcfgs, zhao-titov-2020-visually, zhang-etal-2021-video,  chiu-rush-2020-scaling, chiu2021low, kim2021sequence}.  We use the neural parameterization of \citet{yang-etal-2021-pcfgs} with slight modifications. We show the details in Appd. \ref{appd:pcfg} and Appd. \ref{appd:hmm}.

\section{Experiments}
\subsection{Unsupervised parsing with PCFGs}
\paragraph{Setting.} We evaluate our model on Penn Treebank (PTB)~\citep{marcus-etal-1994-penn}. 
Our implementation is based on the open-sourced code of \citet{yang-etal-2021-pcfgs}\footnote{\url{github.com/sustcsonglin/TN-PCFG}} and we use the same setting as theirs. For all experiments, we set the ratio of nonterminal number to the preterminal number to 1:2 \footnote{Although we did not explicitly distinguish between nonterminal and preterminal symbols previously, in our implementation, we follow \citet{kim-etal-2019-compound} to make such distinction, in which terminal words can only be generated by preterminal symbols, and binary rules can only be invoked by nonterminal symbols.} which is the common practise. We set the rank size to 1000. We show other details in Appd. \ref{appd:data} and \ref{appd:exp}. 


\begin{table}[tb!]
    \centering 
    \scalebox{0.8}{
    \begin{tabular}{lc}
        \toprule 
        {\bf Model} &   {\bf S-F1}  \\
        \midrule
        N-PCFG \cite{kim-etal-2019-compound} & 50.8 \\
        C-PCFG \cite{kim-etal-2019-compound} & 55.2\\
        NL-PCFG \cite{zhu-etal-2020-return}  & 55.3 \\
        TN-PCFG \cite{yang-etal-2021-pcfgs} & 57.7 \\
        NBL-PCFG \cite{yang-etal-2021-neural} & 60.4 \\ 
        \midrule
        Ours with 9000 PTs and 4500 NTs & \textbf{64.1} \\
        \midrule 
            \multicolumn{2}{c}{For reference}\\
        \midrule
        Constituency test \cite{cao-etal-2020-unsupervised} & 62.8 \\
        S-DIORA \cite{drozdov-etal-2020-unsupervised} & 57.6\\
        StructFormer \cite{shen-etal-2021-structformer} & 54.0 \\
        DIORA+span constraint \cite{xu-etal-2021-improved} & 61.2 \\
        \midrule
       \bottomrule 
    \end{tabular}}
    \caption{Results on PTB. S-F1: sentence-level F1. PTs: preterminals. NTs: nonterminals.}
    \label{tab:ptb}
\end{table}

\paragraph{Main result.} Table \ref{tab:ptb} shows the result on PTB. Among previous unsupervised PCFG models, TN-PCFG \cite{yang-etal-2021-pcfgs} uses the largest number of states (500 perterminals and 250 nonterminals). Our model is able to use much more states thanks to our new inside algorithm with lower time complexity, surpassing all previous PCFG-based models by a large margin and achieving a new state-of-the-art in unsupervised constituency parsing in terms of sentence-level F1 score on PTB.

\begin{figure}[tb!]
\centering
\begin{subfigure}{\linewidth}
    \scalebox{0.65}{
\begin{tikzpicture}
\begin{axis}[ ylabel={F1 score $(\/ \%)$}, xlabel={Number of preterminal symbols $(K)$}]
\addplot table[x=x,y=y] {f1.dat};
\addplot [name path=upper,draw=none] table[x=x,y expr=\thisrow{y}+\thisrow{err}] {f1.dat};
\addplot [name path=lower,draw=none] table[x=x,y expr=\thisrow{y}-\thisrow{err}] {f1.dat};
\addplot [fill=blue!10] fill between[of=upper and lower];
\end{axis}
\end{tikzpicture}
}
\caption{}
\end{subfigure}
\begin{subfigure}{\linewidth}
    \scalebox{0.65}{
\begin{tikzpicture}
\begin{axis}[ ylabel={Perplexity}, xlabel={Number of preterminal symbols $(K)$}]
\addplot table[x=x,y=y] {ppl.dat};
\addplot [name path=upper,draw=none] table[x=x,y expr=\thisrow{y}+\thisrow{err}] {ppl.dat};
\addplot [name path=lower,draw=none] table[x=x,y expr=\thisrow{y}-\thisrow{err}] {ppl.dat};
\addplot [fill=blue!10] fill between[of=upper and lower];
\end{axis}
\end{tikzpicture}
}
\caption{}
\end{subfigure}
    \caption{The change of F1 scores and perplexities with the change of number of perterminal symbols.}
    \label{fig:ppl}
\end{figure}
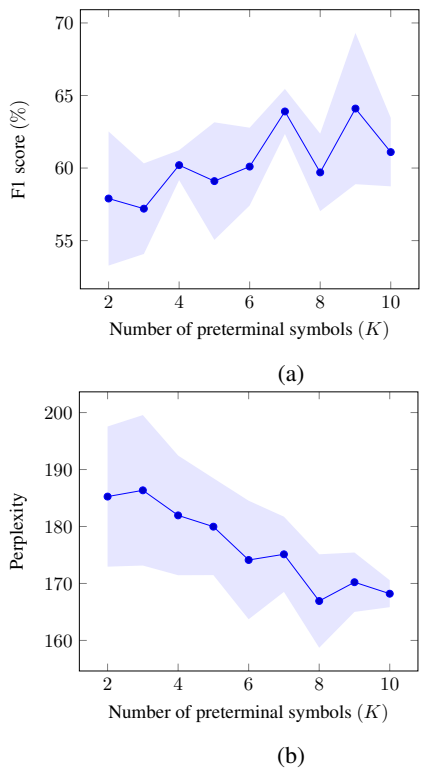

\paragraph{Ablation study.} Fig. \ref{fig:ppl} shows the change of the sentence-level F1 scores and perplexity with the change of the number of preterminals.
As we can see, when increasing the state, the perplexity tends to decrease while
the F1 score tends to increase, validating the effectiveness of using large state spaces for neural PCFG induction.

\subsection{HMM language modeling}
 \paragraph{Setting.} We conduct the language modeling experiment also on PTB. 
Our implementation is based on the open-sourced code of \citet{chiu2021low}\footnote{\url{github.com/justinchiu/low-rank-models}}. We set the rank size to 4096. See Appd. \ref{appd:data} and \ref{appd:exp} for more details.

\paragraph{Main result.}
Table \ref{tab:ptb_lm} shows the perplexity on the PTB validation and test sets. As discussed earlier, VL-HMM \cite{chiu-rush-2020-scaling} imposes strong sparsity constraint to decrease the time complexity of the forward algorithm and requires pre-clustering of terminal symbols. Specifically, VL-HMM uses Brown clustering \cite{brown-etal-1992-class}, introducing external information to improve performance. Replacing Brown clustering with uniform clustering leads to a 10 point increase in perplexity on the PTB validation set. LHMM \cite{chiu2021low} and our model only impose low-rank constraint without using any external information and are thus more comparable. Our method outperforms LHMM by $4.8$ point when using the same state number (i.e., $2^{14}$), and it can use more states thanks to our lower inference time complexity.

\begin{table}[tb!]
    \centering 
    \scalebox{0.75}{
    \begin{tabular}{lcc}
        \toprule 
        {\bf Model} &  {\bf Val} &  {\bf Test}  \\
                \midrule
        VL-HMM ($2^{15}$ states, Brown) & 125.0 & 116.0\\
        VL-HMM ($2^{14}$ states, Brown)$^\dagger$ & 136 & -\\
        VL-HMM ($2^{14}$ states, Uniform)$^\dagger$  & 146 & - \\  
        LHMM ($2^{14}$ states) & 141.4 & 131.8 \\
        \midrule
        Ours ($2^{14}$ states) & 135.6 & 127.0\\ 
        Ours ($2^{15}$ states) & 137.0 & 126.4 \\
        \midrule 
        \multicolumn{2}{c}{For reference}\\
        \midrule
        HMM+RNN \cite{Buys2018BridgingHA}  & 142.3 & - \\
        AWD-LSTM \cite{DBLP:conf/iclr/MerityKS18} & 60.0 & 57.3 \\ 
        \midrule
        \midrule
       \bottomrule 
    \end{tabular}}
    \caption{Resulting perplexity on PTB validate set and test set. VL-HMM: \cite{chiu-rush-2020-scaling}. LHMM: \cite{chiu2021low}.
    $\dagger$ denotes results reported by ablation study of \citet{chiu-rush-2020-scaling}.}
    \label{tab:ptb_lm}
\end{table}

\paragraph{Ablation study.} 
As we can see in Table \ref{tab:lm_ablation}, the perplexity tends to decrease when increasing the state number, validating the effectiveness of using more states for neural HMM language modeling.

\begin{table}[tb!]
    \centering 
    \scalebox{0.8}{
    \begin{tabular}{lcc}
        \toprule 
        {\bf \#States} &  {\bf Val} &  {\bf Test}  \\
                \midrule
        $2^{12}$ & 149.8 & 139.1\\
        $2^{13}$ & 143.8 & 133.4\\
        $2^{14}$ & 149.5 & 137.4 \\  
        $2^{15}$ & 141.1 & 131.1 \\
       \bottomrule 
    \end{tabular}}
    \caption{Perplexity with varying numbers of states. Following \citet{chiu2021low}, we fix the rank to 2048 for faster ablation studies.}
    \label{tab:lm_ablation}
\end{table}

\paragraph{Discussion.} It is interesting to note that our HMM model is roughly equivalent to another HMM with interchanged rank and state sizes as can be seen in Fig.\ref{fig:new_hmm}(c). To verify this equivalence, we run LHMM in the original state space with 2048 states and rank $2^{15}$. The resulting perplexity is 133.49 on average on the PTB test set, which is worse than that of ours (126.4). 
We leave further experimentation and analyses of this discrepancy for future work.

\section{Related work}
Tensor and matrix decomposition have been used to decrease time and space complexities of probabilistic inference algorithms.  \citet{DBLP:journals/jmlr/SiddiqiBG10} propose a reduced-rank HMM whereby the forward algorithm can be carried out in the rank space, which is similar to our model, but our method is more general.
\citet{DBLP:conf/nips/CohenC12, cohen-etal-2013-approximate} use CPD for fast (latent-variable) PCFG parsing, but they do not leverage CPD for fast learning and they need to actually perform CPD on existing probability tensors. \citet{DBLP:conf/aistats/RabusseauBC16} use low-rank approximation method to learn weighted tree automata, which subsumes PCFGs and latent-variable PCFGs. Our method can subsume more models.
\citet{yang-etal-2021-pcfgs, yang-etal-2021-neural} propose CPD-based neural parameterizations for (lexicalized) PCFGs. \citet{yang-etal-2021-pcfgs} aim at scaling PCFG inference. We achieve better time complexity than theirs and hence can use much more hidden states. \citet{yang-etal-2021-neural} aims to decrease the complexity of lexicalized PCFG parsing, which can also be described within our framework. 
\citet{chiu2021low} use low-rank matrix decomposition, which can be viewed as CPD on order-2 tensors, to accelerate inference on chain and tree structure models including HMMs and PCFGs. However, their method is only efficient when the parameter tensors are of order 2, e.g., in HMMs and HSMMs. Our method leverages full CPD, thus enabling efficient inference with higher-order factors, e.g., in PCFGs. Our method can be applied to all models considered by \citet{chiu2021low}, performing inference in the rank-space with lower complexities. 

Besides HMMs and PCFGs,  \citet{DBLP:conf/icml/WrigleyLY17} propose an efficient sampling-based junction-tree algorithm using CPD to decompose high-order factors. \citet{DBLP:journals/corr/abs-2010-09283} also use CPD to decompose high-order factors for fast belief propagation. \citet{yang2022modeling} use CPD to decompose second-order factors in semantic dependency parsing to  accelerate second-order parsing with mean-field inference. Besides CPD, 
\citet{DBLP:conf/pgm/DucampBNW20} use tensor train decomposition for fast and scalable message passing in Bayesian networks. \citet{JMLR:v22:18-431} leverage matrix product states (i.e., tensor trains) for scalable discrete probabilistic inference. \citet{DBLP:conf/aistats/MillerRT21} leverage tensor networks for fast sequential probabilistic inference.



\section{Conclusion and future work}
In this work, we leveraged tensor rank decomposition (CPD) for low-rank scaling of structured inference. We showed that CPD amounts to decomposing a large factor into several smaller factors connected by a new rank node, and gave a unifying perspective towards previous low-rank structured models \cite{yang-etal-2021-pcfgs, chiu2021low}.  We also presented a novel framework to design a family of rank-space inference algorithms for B-FGGs, a subset of FGGs which subsume most structured models of interest to the NLP community. We have shown the application of our method in scaling PCFG and HMM inference, and experiments on unsupervised parsing and language modeling validate the effectiveness of using large state spaces facilitated by our method. 

We believe our framework can be applied to many other models which have high inference time complexity and are subsumed by B-FGGs, including lexicalized PCFGs, quasi-synchronous context-free grammars (QCFGs), etc. A direct application of our method is to decrease the   inference complexity of the neural QCFG \cite{kim2021sequence}. 

\section*{Acknowledgments}
This work was supported by the National Natural Science Foundation of China (61976139).

\bibliography{anthology,custom}
\bibliographystyle{acl_natbib}

\appendix

\section{Neural parameterization of PCFGs}
\label{appd:pcfg}
In this section, we give the full parameterization of PCFGs. We follow \citet{yang-etal-2021-pcfgs} with slight modifications for generations of $\mathbf{U}, \mathbf{V}$, $\mathbf{W} \in \mathbb{R}^{m\times r}$ in \ref{sec:rank-inside}. We use the same MLPs with two residual layers as \citet{yang-etal-2021-pcfgs}:
\begin{align*}
    \mathbf{s} &= \frac{\exp(\mathbf{u}_S^Th_1(\mathbf{w}_A)}{\sum_{A' \in \mathcal{N}}\exp(\mathbf{u}_S^Th_1(\mathbf{w}_{A'}))}\\
    \mathbf{E} &= \frac{\exp(\mathbf{u}_E^Th_2(\mathbf{w}_t)}{\sum_{E' \in \Sigma}\exp(\mathbf{u}_{E'}^Th_2(\mathbf{w}_{t}))}\\
    \mathbf{U} &= \frac{\exp(\mathbf{u}_H^Tf_1(\mathbf{w}_n)}{\sum_{n' \in \mathcal{N}}\exp(\mathbf{u}_H^Tf_1(\mathbf{w}_{n'}))}\\
    \mathbf{V} &= \frac{\exp(\mathbf{u}_H^Tf_2(\mathbf{w}_l)}{\sum_{H' \in \mathcal{H}}\exp(\mathbf{u}_{H'}^Tf_2(\mathbf{w}_{l}))}\\
    \mathbf{W} &= \frac{\exp(\mathbf{u}_H^Tf_3(\mathbf{w}_l)}{\sum_{H' \in \mathcal{H}}\exp(\mathbf{u}_{H'}^Tf_3(\mathbf{w}_{l}))}\\
    h_i(\mathbf{x}) &= g_{i,1}(g_{i,2}(\tilde{\mathbf{W}_i}\mathbf{x}))\\
    g_{i,j}(\mathbf{y}) &= ReLU(\tilde{\mathbf{V}}_{i,j}ReLU(\tilde{\mathbf{U}}_{i,j}\mathbf{y})) + \mathbf{y}\\
\end{align*}

where $\Sigma$ is the vocabulary set, $\mathcal{H}$ is the set of rank, $\mathcal{N}$ is a finite set of nonterminals, $\mathbf{W}_l = [\mathbf{W}_n;\mathbf{W}_t], \mathbf{w}_l, \mathbf{w}_n, \mathbf{w}_t \in \mathbf{W}_l, \mathbf{W}_n, \mathbf{W}_t$. The main differences of neural parameterization between ours and previous work are that we make the projection parameter $\mathbf{u}_H$ shared among $\mathbf{U}, \mathbf{V},\text{ and } \mathbf{U}$.

\section{Neural parameterization of HMMs}
\label{appd:hmm}
In this section, we give the full parameterization of HMMs, which is similar to PCFGs' parameterization. Define $\mathbf{s}$ as start probability for HMMs. And the definitions of $\mathbf{U}, \mathbf{V}, \mathbf{W}$ are same as definitions in $\ref{sec:rank-forward}$:
\begin{align*}
    \mathbf{s} &= \frac{\exp(\mathbf{u}_P^Th_1(\mathbf{w}_s))}{\sum_{s' \in \mathcal{S}}\exp(\mathbf{u}_P^Th_1(\mathbf{w}_{s'}))}\\
    \mathbf{U} &= \frac{\exp(\mathbf{u}_H^T\mathbf{w}_u)}{\sum_{H' \in \mathcal{H}}\exp(\mathbf{u}_{H'}^T\mathbf{w}_u)}\\
    \mathbf{V} &= \frac{\exp(\mathbf{u}_H^T\mathbf{w}_v)}{\sum_{v' \in \mathcal{S}}\exp(\mathbf{u}_{H}^T\mathbf{w}_{v'})}\\
    \mathbf{W} &= \frac{\exp(\mathbf{u}_W^Th_2(\mathbf{w}_w)}{\sum_{w' \in \Sigma}\exp(\mathbf{u}_W^Th_2(\mathbf{w}_{w'}))}\\
    h_i(\mathbf{x}) &= g_{i,1}(g_{i,2}(\tilde{\mathbf{W}_i}\mathbf{x}))\\
    g_{i,j}(\mathbf{y}) &= ReLU(\tilde{\mathbf{V}}_{i,j}ReLU(\tilde{\mathbf{U}}_{i,j}\mathbf{y})) + \mathbf{y}\\
\end{align*}

where $\mathcal{S}$ is a finite set of states, $\mathcal{H}$ is the set of rank, $\Sigma$ is vocabulary set.

\section{Data details}
\label{appd:data}
Penn Treebank (PTB) \citep{marcus-etal-1994-penn}\footnote{The licence of PTB dataset is LDC User Agreement for Non-Members, which can be seen on \url{https://catalog.ldc.upenn.edu/LDC99T42}} consists of 929k training words, 73k validation words, and 82k test words, with a vocabulary of size 10k. 

For PCFGs, we follow \citet{yang-etal-2021-pcfgs} and use their code to preprocess dataset. This processing discards punctuation and lowercases all tokens with 10k most frequent words as the vocabulary. The splits of the dataset are: 2-21 for training, 22 for validation and 23 for test.

For HMMs, we follow \citet{chiu2021low} and use their code to preprocess dataset. We lowercase all words and substitutes OOV words with UNKs. EOS tokens have been inserted after each sentence.

\section{Experimental details}
\label{appd:exp}
For PCFGs, we use Xavier normal initialization to initialize the weights in $h_i$ and $f_i$. We optimize our model using Adam optimizer with $\beta_1 = 0.75, \beta_2 = 0.999$, and the learning rate $0.002$, setting the dimension of all embeddings to $256$. 

For HMMs, we initialize all parameters by Xavier normal initialization except for $\mathbf{w}_s$ and $\mathbf{w}_w$. We use AdamW optimizer with $\beta_1 = 0.99, \beta_2 = 0.999$, and the learning rate $0.001$, and a max grad norm of $5$. We use dropout rate of $0.1$ to dropout $\mathbf{w}_s$ and $\mathbf{U},\mathbf{V}$ in HMMs. We train for 30 epochs with a max batch size of 256 tokens, and reduce the learning by multiplying $\frac{1}{2}$ if the validation perplexity fails to improve after 2 evaluations. Evaluations are performed one time per epoch. We follow \citet{chiu2021low} to shuffle sentences and leverage bucket iterator, where batch of sentences are drawn from buckets containing sentences of similar lengths to minizing padding.

We run all experiments on NVIDIA TITAN RTX and NVIDIA RTX 2080ti and all experimental results are averaged from four runs.

\end{document}